\documentclass[10pt,twocolumn,letterpaper]{article}

\usepackage{wacv}
\usepackage{times}
\usepackage{epsfig}
\usepackage{graphicx}
\usepackage{amsmath}
\usepackage{amssymb}
\usepackage[accsupp]{axessibility}  
\usepackage[utf8]{inputenc}
\usepackage[T1]{fontenc}
\usepackage{textcomp}
\usepackage{multirow}
\usepackage{booktabs}
\usepackage{colortbl}
\usepackage{enumitem}
\usepackage{algorithm}
\usepackage{algpseudocode}
\usepackage{listings}
\usepackage{tablefootnote}

\usepackage[flushleft]{threeparttable}
\definecolor{mygray}{gray}{.9}
\usepackage[accsupp]{axessibility}

\def\wacvPaperID{****} 

\wacvfinalcopy 

\ifwacvfinal
\fi

\ifwacvfinal
\usepackage[breaklinks=true,bookmarks=false]{hyperref}
\else
\usepackage[pagebackref=true,breaklinks=true,colorlinks,bookmarks=false]{hyperref}
\fi

\begin{document}

\title{MTGLS: Multi-Task Gaze Estimation with Limited Supervision}

\author{Shreya Ghosh$^{1}$ Munawar Hayat$^{1}$ Abhinav Dhall$^{1,2}$ Jarrod Knibbe$^{3}$ \\
$^{1}$Monash University $^{2}$Indian Institute of Technology Ropar $^{3}$University of Melbourne\\
{\tt\small \{shreya.ghosh, munawar.hayat, abhinav.dhall\}@monash.edu jarrod.knibbe@unimelb.edu.au }}


\maketitle

\ifwacvfinal
\thispagestyle{empty}
\fi

\begin{abstract}
 Robust gaze estimation is a challenging task, even for deep CNNs, due to the non-availability of large-scale labeled data. Moreover, gaze annotation is a time-consuming process and requires specialized hardware setups. We propose MTGLS: a Multi-Task Gaze estimation framework with Limited Supervision, which leverages abundantly available non-annotated facial image data.\ MTGLS distills knowledge from off-the-shelf facial image analysis models, and learns strong feature representations of human eyes, guided by three complementary auxiliary signals: (a) the line of sight of the pupil (i.e. pseudo-gaze) defined by the localized facial landmarks, (b) the head-pose given by Euler angles, and (c) the orientation of the eye patch (left/right eye). To overcome inherent noise in the supervisory signals, MTGLS further incorporates a noise distribution modelling approach. Our experimental results show that MTGLS learns highly generalized representations which consistently perform well on a range of datasets. Our proposed framework outperforms the unsupervised state-of-the-art on CAVE (by $\sim 6.43$\%) and even supervised state-of-the-art methods on Gaze360 (by $\sim 6.59\%$) datasets.  
\end{abstract}

\section{Introduction}
Eye movement provides important cues for nonverbal behaviour analysis, revealing insights into visual attention~\cite{liu2011visual} and human cognition (emotions, beliefs, and desires)~\cite{wang2017deep}. Such insights have proved valuable across multiple application scenarios, including gaming platforms~\cite{cheng2013gaze}, measuring student's engagement~\cite{mustafa2018prediction}, determining driver's attention~\cite{ghosh2021speak2label}, etc. Recently, significant research progress has been made across a broad range of gaze related tasks, including gaze tracking~\cite{wang2019neuro,sun2014toward,park2018deep,gaze360_2019}, pupil detection~\cite{tonsen2016labelled}, pupil and iris segmentation~\cite{varkarakis2020deep}, and gaze estimation~\cite{park2018deep,park2019few,gaze360_2019,yu2019improving,zhang2017mpiigaze}. Automatic gaze estimation still remains challenging due to the individuality of eyes, eye-head interplay, occlusion, image quality, and illumination conditions. The current state-of-the-art gaze estimation methods~\cite{park2019few,yu2019unsupervised,yu2019improving,park2018deep} employ deep Convolution Neural Networks (CNNs) to address these challenges. They generally require a lot of annotated data and person-specific calibration. 
Accurate annotation of gaze data is a complex, noisy, resource expensive, and time-consuming task. The annotation process can easily become noisy due to participants' non-cooperation (e.g., distraction), involuntary actions (e.g., eye blink), and measurement errors introduced by different data acquisition setups which limits merging multiple datasets. Research has sought to address these annotation challenges through data generation (e.g., synthesizing additional labelled data via gaze redirection~\cite{yu2019unsupervised} and GANs~\cite{wang2018hierarchical,sugano2014learning}). However, these complex models are still constrained by the high computational complexity and quality of synthesized samples. 

To this end, weakly/Semi/Self-Supervised Learning provides a promising paradigm as it enables learning from a large amount of readily available data without requiring any manual annotation efforts. To date, there are only a few existing works~\cite{yu2019unsupervised,dubey2019unsupervised} that learn low dimensional gaze representation guided by supervisory signals in the form of data augmentation, gaze redirection, and gaze zone. These existing works are either guided by weak auxiliary signals (e.g., gaze zone~\cite{dubey2019unsupervised}), or are hard to scale since they require the generation of the eye or face image~\cite{yu2019unsupervised}. 
\begin{figure}[t]
    \centering
    \includegraphics[width=0.94\linewidth]{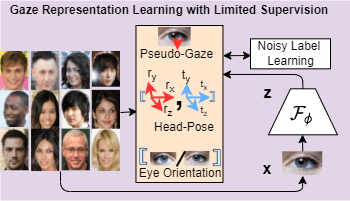}
    \caption{Overview of the proposed Multi-Task Gaze estimation framework with Limited Supervision (MTGLS). MTGLS learns a meaningful gaze embedding space guided by different auxiliary signals from `in-the-wild' facial images curated from the web. }
\label{fig:eye_teaser}
\vspace{-5mm}
\end{figure}
To learn a generalized and compact gaze representation, we propose an framework with a multi-task learning objective. Our representation learning benefits from the progress made across other facial image analysis tasks (i.e. head-pose and facial landmarks), and is guided by three auxiliary gaze-relevant signals: (a) Line Of Sight (LOS) of the pupil (pseudo-gaze) determined by automatically localized facial landmarks~\cite{barz2020deep}, (b) head-pose in terms of Euler angles given by~\cite{albiero2020img2pose}, and (c) orientation of eye patch (left/right eye). These auxiliary signals provide complementary supervision to MTGLS. Our learned representations are highly generic and transferable. Unlike the existing methods~\cite{park2019few,yu2019improving}, which require person-specific calibration to adapt the label space, our approach achieves competitive results without such calibration, which is highly desirable in wearable eye devices, especially in augmented and virtual reality glasses~\cite{swaminathan2018enabling,blattgerste2018advantages}. Moreover, our extensive empirical evaluations demonstrate that the representations learned by MTGLS generalize well across a range of datasets for different eye gaze estimation tasks. We further show that MTGLS can be used for real-time gaze inference applications, especially for driver's engagement. The \textbf{main contributions} of the paper are: 

\begin{itemize}[topsep=1pt,itemsep=0pt,partopsep=1ex,parsep=1ex,leftmargin=*]
    \item MTGLS, a novel multi-task gaze estimation framework with limited supervision for rich visual representation learning, guided by complementary information in the form of line-of-sight of the pupil, head-pose, and orientation of eye patch. 
    \item To overcome the impact of inherent noise in the auxiliary label space, we incorporate a probability distribution modelling based noise correction strategy which further boosts the performance of the model.
    \item We demonstrate the effectiveness of the proposed approach by extensive evaluations on downstream gaze estimation tasks on four benchmark datasets (i.e. CAVE, MPII, Gaze360, and DGW). The proposed approach not only outperforms unsupervised state-of-the-art methods on CAVE (3.42\textdegree\ \textrightarrow 3.20\textdegree, $\sim 6.43\%$), but also performs favorably on the supervised state-of-the-art method on Gaze360 (13.80\textdegree\ \textrightarrow 12.89\textdegree, $\sim 6.59\%$) datasets. 
\end{itemize}

\section{Related Work}
\label{sec:prior}

\begin{figure*}
    \centering
    \includegraphics[width=\linewidth]{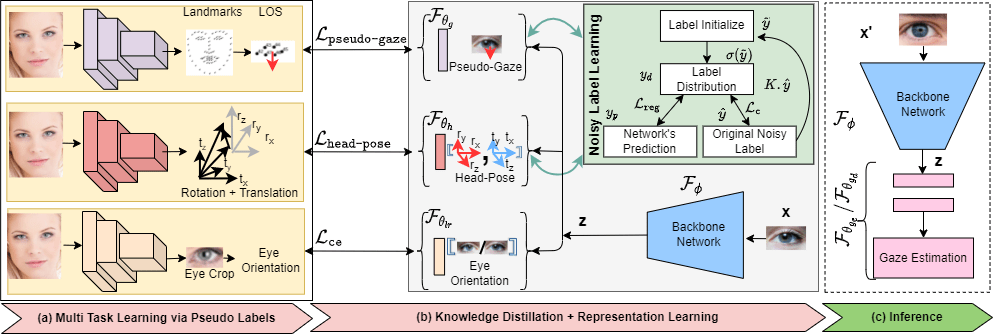}
    \caption{An overview of our proposed MTGLS framework. From left to right, we show \textit{(a) Multi-task Learning via Pseudo Labels:} MTGLS learns gaze representations with limited supervision, by distilling knowledge from off-the-shelf deep models developed for facial image modelling. Here, MTGLS utilizes LOS, head-pose and eye orientation as auxiliary signals (Refer Sec.~\ref{sec:auxiliary}); \textit{(b) Knowledge Distillation and Representation Learning:} Our gaze estimation framework consists of the backbone network and noisy label learning module (Refer Sec.~\ref{sec: NLL}). During the representation learning process, MTGLS considers the impact of inherent noise in the assigned soft label space; \textit{(c) Inference:} For downstream adaptation, we use Linear Probing (LP), Fine-Tuning (FT) to show the generalization and transferability of the learnt features across different tasks.}
    \label{fig:pipeline}
    \vspace{-5mm}
\end{figure*}

\noindent \textbf{Gaze Estimation.} 
A thorough analysis of gaze estimation literature is mentioned in a recent survey~\cite{ghosh2021Automatic}. Traditionally, gaze estimation methods are based on appearance features~\cite{morimoto2000pupil,zhu2002subpixel,hennessey2006single}, or geometric attributes such as iris and corner detection~\cite{torricelli2008neural}. Mostly different facial regions are widely used as input Region of Interest (RoI) in the literature such as whole face~\cite{cheng2020coarse,dubey2019unsupervised,zhang2017s,park2019few}, eye region~\cite{jyoti2018automatic,zhang2015appearance,park2018learning,park2018deep}. These features are then mapped to the screen coordinates using a learning algorithm including support vector regression~\cite{smith2013gaze} or deep neural nets (DNNs) models~\cite{krafka2016eye,zhang2015appearance,zhang2017s,zhang2017mpiigaze,jyoti2018automatic,FischerECCV2018}. Among the DNNs based methods, supervised learning methods~\cite{zhang2015appearance,jyoti2018automatic,zhang2017mpiigaze} have been most popular and successful, but they require a large amount of annotated training data. 
Some studies~\cite{sugano2014learning,benfold2011unsupervised,zhang2017everyday,santini2017calibme,park2019few,karessli2017gaze,he2019device} explore gaze estimation in unsupervised and weakly/self/semi-supervised settings to reduce the data annotation burden. These approaches are mainly based on conditional random field~\cite{benfold2011unsupervised}, unsupervised gaze target discovery~\cite{zhang2017everyday}, few-shot learning~\cite{park2019few,yu2019improving}, weakly-supervised~\cite{kothari2021weakly}, and representation learning~\cite{yu2019unsupervised,dubey2019unsupervised}. Apart from these methods, Bayesian learning paradigms are also explored in the context of generalization~\cite{wang2019generalizing,wang2018hierarchical}. Nowadays there is an increasing trend to inject domain knowledge in terms of certain heuristics for gaze estimation such as the correlation between head-pose and gaze direction~\cite{park2019few,zheng2020self}, two eye symmetry~\cite{cheng2020gaze}, facial landmark based gaze modelling~\cite{wang2019generalizing} and uncertainty in gaze modelling~\cite{kothari2021weakly,gaze360_2019}.


\noindent \textbf{Gaze Datasets.} The performance of existing gaze estimation methods is generally dependent upon the amount of annotated data used during training. While several efforts have been made to collect and annotate gaze data in the previous decade~\cite{smith2013gaze,FunesMora_ETRA_2014,huang2015tabletgaze,zhang2017mpiigaze,zhang2017s,FischerECCV2018,gaze360_2019,garbin2019openeds,palmero2020openeds2020,zhang2020eth,Park2020ECCV}, gaze data collection has multiple challenges, including the requirement of specialized hardware setup, inconsistency amongst different acquisition apparatus which limits merging multiple datasets, and resource-expensive data annotation process. Some approaches turn towards synthetic data generation to increment the size of the training data for gaze estimation task~\cite{ganin2016deepwarp,he2019photo,wood2018gazedirector,yu2019improving,sela2017gazegan}. 
While these methods can enhance the training set size, the quality of their synthesized samples differ from their original counterparts. 
More specifically, these methods inherit several domain gaps, including content preservation (e.g., eye shape, eye appearance), network complexity, and resource expensiveness. Additionally, the gaze re-direction methods may have limited gaze direction samples while extrapolating on a new subject. These domain gaps are quite challenging and difficult to eliminate. While gaze estimation still remains a challenging feat, limited by the capacity of available annotated data to train supervised models, significant research progress has been made across other facial analysis tasks, including face identification, classification, landmark localization, and head-pose estimation. 
In this work, we propose to leverage from existing facial image analysis models and distill their knowledge for gaze estimation, by making use of readily available facial image data. We demonstrate that the gaze representation learned by our method generalizes well for gaze estimation tasks across a range of datasets~\cite{zhang2017mpiigaze,smith2013gaze,gaze360_2019,ghosh2021speak2label}, including those collected in controlled environments~\cite{smith2013gaze}, and more challenging unconstrained datasets~\cite{zhang2017mpiigaze,gaze360_2019}. Moreover, owing to the strong generalization capabilities of our learned representations, MTGLS can align the label space manifold of auxiliary and downstream gaze estimation label space which consequently helps in person-specific adaptation. Thus, it is quite promising for gaze calibration tasks having potential applications in AR/VR glasses where unsupervised gaze calibration is highly desirable. \textit{Please note that the proposed method can be considered as self-supervised. We have used the term limited supervision as the pseudo labels are obtained from off-the-shelf models~\cite{wang2021facex,albiero2020img2pose}.}

\section{Method} \label{sec:method}
Human eye gaze depends on several factors, including the relative location of the pupils, head-pose, and appearance. A strong feature representation of the eye is crucial for robust gaze modelling. Instead of limiting ourselves to scarcely available gaze annotated data, we distill knowledge from a vast amount of non-annotated data of human faces, which is easily available on the web. Our proposed framework, MTGLS is guided by three auxiliary signals: LOS of the pupil (pseudo-gaze), head-pose, and orientation of eye patch (i.e. if the eye patch belongs to the left or right eye). The proposed multi-objective learning formulation maps the input eye patch image to the optimal output representation manifold, which can further be utilized for other gaze estimation tasks. The visual illustration of MTGLS is shown in Fig.~\ref{fig:pipeline} and an algorithm is shown in Alg.~\ref{alg:MTGLS}.

\subsection{Gaze Representation Learning}

\noindent \textbf{Preliminaries.}
Lets assume our MTGLS model $\mathcal{F}$ consists of feature embedding parameter $\phi$ and label space parameters $\theta$. An input image $\mathbf{x}$ is mapped to the feature embedding space by $\mathcal{F}_{\phi}:\mathbf{x} \to \mathbf{z}$, where $ \mathbf{z}\in \mathbb{R}^d $. Consequently, the features $\mathbf{z}$ are mapped to three auxiliary output spaces $ \mathbf{g}\in \mathbb{R}^3 $, $ \mathbf{h}\in \mathbb{R}^6 $, and $ \mathbf{lr}\in \mathbb{R}^2 $, respectively by three branches: $\mathcal{F}_{\theta_g}:\mathbf{z} \to \mathbf{g}$, $\mathcal{F}_{\theta_h}:\mathbf{z} \to \mathbf{h}$, and $\mathcal{F}_{\theta_{lr}}:\mathbf{z} \to \mathbf{lr}$. Here, $\mathbf{g}$, $\mathbf{h}$, and $\mathbf{lr}$ are the pseudo-gaze, head-pose, and eye orientation, respectively. The model $\mathcal{F}$ therefore consists of joint learning of three sub-networks: $(\mathcal{F}_\phi \circ \mathcal{F}_{\theta_g})$, $(\mathcal{F}_\phi \circ \mathcal{F}_{\theta_h})$, and $(\mathcal{F}_\phi \circ \mathcal{F}_{\theta_{lr}})$. MTGLS is trained on dataset $\mathcal{D}$ which consists of $n$ image-label pairs denoted as: $\{\mathbf{x_i},\mathbf{y_i}\}_n$, where $\mathbf{y_i}=\{ \mathbf{g_i}, \mathbf{h_i}, \mathbf{lr_i}\}$ are the auxiliary task pseudo-labels, estimated by off-the-shelf facial image analysis models. The learned features by $\mathcal{F}$ are transferred and adapted for two gaze estimation tasks i.e. $\mathcal{F}_\phi  \circ \mathcal{F}_{\theta_{g_{e}}}$ for gaze estimation and $\mathcal{F}_\phi  \circ \mathcal{F}_{\theta_{g_{d}}}$ for driver's gaze zone estimation, using separate tasks specific modules i.e.,  $\mathcal{F}_{\theta_{g_{e}}}$ and $\mathcal{F}_{\theta_{g_{d}}}$.

\subsubsection{Auxiliary Tasks} \label{sec:auxiliary}
Self-training via knowledge distillation aims to learn rich representative features from a small initial annotated set to a large scale unlabeled instances with soft label assignment~\cite{xie2020self}. Knowledge distillation~\cite{hinton2015distilling} is widely used in literature for representation learning~\cite{caron2021emerging,georgescu2021anomaly,grill2020bootstrap}. MTGLS also adopt this soft label assignment strategy on large scale non-annotated data. Several prior works have reported the performance enhancement with the combination of representation learning and knowledge distillation~\cite{fang2021seed,noroozi2018boosting}. Along these lines, we propose the following auxiliary tasks aligned with gaze estimation for limited supervision.

\noindent \textbf{a) Pseudo-Gaze:}
Prior literature on unsupervised and weakly supervised gaze representation learning utilize sparse gaze zone~\cite{dubey2019unsupervised}, gaze redirection~\cite{yu2019unsupervised} and looking at each other~\cite{kothari2021weakly} soft labelling strategies for limited-supervision. Similarly, we define landmark-based LOS of the pupil as an auxiliary task and term it as \textit{pseudo-gaze}. First, the eye region is registered via FaceX-Zoo's state-of-the-art landmark detection method~\cite{wang2021facex}. We utilize the detected pupil center and eye landmarks to compute the gaze vector. The estimated gaze vector is from the 3-D eyeball center to the pupil location where the eyeball center is approximated by the eye landmarks. We follow the conventional details of this approximation technique from prior works~\cite{park2018learning,8373812}. We use the cosine similarity based gaze direction loss (similar to the standard loss used in this domain~\cite{zhang2020eth}) to train the $(\mathcal{F}_\phi \circ \mathcal{F}_{\theta_g})$ sub-network on pseudo-gaze labels, defined by, $\mathcal{L}_{\texttt{pseudo-gaze}} = \frac{\mathbf{g}}{||\mathbf{g}||_2}. \frac{\mathbf{g'}}{||\mathbf{g'}||_2},$ where, $\mathbf{g}$ and $\mathbf{g'}$ respectively denote the pseudo-gaze (ground truth) and predicted gaze in auxiliary label space.

\noindent \textbf{b) Head-Pose:}
The head-pose also carries important information for gaze estimation, since there exists a strong correlation between eye and head once a person looks at something~\cite{zhang2015appearance,park2019few,zheng2020self}. We use the state-of-the-art head-pose estimator `\texttt{img2pose}'~\cite{albiero2020img2pose} to infer pseudo 6-D head-pose labels from the input face (Refer Fig.~\ref{fig:pipeline} (a)). Given a face image $\mathbf{x}$, we estimate 6-DoF pose from the \texttt{img2pose} model. We minimize the mean-squared error between predicted head-pose $\mathbf{h'}$, and pseudo head-pose (ground truth) $\mathbf{h}$ to train sub-network $(\mathcal{F}_{\phi} \circ \mathcal{F}_{\theta_h})$. The loss is defined as $\mathcal{L}_{\texttt{head-pose}} = MSE(\mathbf{h},\mathbf{h'})$.

\noindent \textbf{c) The Orientation of Eye Patch:}
We further take benefit from the eye orientation information to strengthen our representation learning. To this end, we define an auxiliary task in terms of eye orientation i.e., whether the eye patch belongs to the left or right eye of a subject. We minimize the binary cross-entropy loss ($\mathcal{L}_{\texttt{ce}}$) for left-right eye patch prediction corresponding to the $(\mathcal{F}_\phi \circ \mathcal{F}_{\theta_{lr}})$ sub-network.

\begin{algorithm}[tb]
\caption{Noisy Label Learning (NLL) for MTGLS}
\label{alg:nll}
\begin{algorithmic}[1]
\Require{$\mathcal{F}_\phi$, $\mathcal{F}_{\theta_g}$, $\mathcal{F}_{\theta_h}$, $\mathbf{Y_d}$, $\mathbf{x}$, $\mathbf{\tilde{y}}$, $\mathbf{\hat{y}}$, $\mathbf{y_d}$, and $\mathbf{y_p}$}
\For{\texttt{$\mathbf{e}$ epochs}} \Comment{Training}
\State $\mathbf{\hat{y}}\gets$ noisy labels
\State $\mathbf{\tilde{y}}\gets K.\mathbf{\hat{y}}$ 
\State $\mathbf{y_d}\gets \sigma (\mathbf{\tilde{y}})$
\State $\mathbf{y_p}\gets \mathcal{F}_\phi \circ \mathcal{F}_{\theta} (\mathbf{x})$ \Comment{$\mathcal{F}_{\theta} \in \{\mathcal{F}_{\theta_g}, \mathcal{F}_{\theta_h} \}$}
\State $\mathcal{L} = \mathcal{L}_{\texttt{reg}} (\mathbf{y_p}, \mathbf{y_d}) + \mathcal{L}_{\texttt{c}} (\mathbf{y_d}, \mathbf{\hat{y}}) + {||\omega||}^{2}_2 $
\State $\{\phi, \theta, \mathbf{Y_d} \}\gets \{\phi, \theta, \mathbf{Y_d} \} - \triangledown_{\{\phi, \theta, \mathbf{Y_d} \}} \mathcal{L}$
\EndFor
\end{algorithmic}
\end{algorithm}

\subsubsection{Noisy Label Learning}
\label{sec: NLL}

Apart from the eye orientation labels, the other auxiliary signals (i.e. pseudo-gaze and headpose labels) are based upon off-the-shelf deep models~\cite{barz2020deep,albiero2020img2pose}, and contain noise. To overcome inherent noise in auxiliary label space, we incorporate Noisy Label Learning (NLL). Our NLL module is inspired by~\cite{yi2019probabilistic} and is summarized in Alg.~\ref{alg:nll}. Let's define $\mathbf{\tilde{y}} = K.\mathbf{\hat{y}}$, where, $\mathbf{\hat{y}}$ are the noisy labels and $K$ is a scalar. For our experiments, we empirically choose $K = 10$ similar to~\cite{yi2019probabilistic}. 
We further define the label distribution $\mathbf{y_d} = \sigma (\mathbf{\tilde{y}})$, where, $\sigma$ is the sigmoid operator. Here, $\mathbf{y_d}$ attempts to model the unknown noise-free label distribution for inputs $\mathbf{X}: \{\mathbf{x_i}, \forall\ 1 \leq i \leq n\}$ via learning. Let, $\mathbf{Y_d}$ be the parameters of the label distribution $\mathbf{y_d}$ defined as $\mathbf{Y_d}=\{\mathbf{Y_d^i},\ \forall\ 1 \leq i \leq n\}$. During each iteration of the training process, the parameters of $\mathbf{Y_d}$ (i.e., $\mathbf{Y_d^i}$) are updated along with the network parameters (i.e. $\phi$ and $\theta$) via back-propagation process~\cite{tanaka2018joint,yi2019probabilistic}. We also hypothesize that $\mathbf{\tilde{y}}$ assists $\mathbf{y_d}$ to be a valid probability distribution. Thus, $ \mathbf{y_d}$ is initialized in terms of $\mathbf{\tilde{y}}$ (i.e. $\sigma (K.\mathbf{\hat{y}})$) instead of random initialization. Hence, after initialization $ \mathbf{y_d} \approx \mathbf{\hat{y}}$.
The overall learning process is guided by three loss functions:

\noindent \textbf{Regression Loss ($\mathcal{L}_{\texttt{reg}}$).} is given by KL divergence between backbone network's prediction ($\mathbf{y_p}$) and label distribution ($\mathbf{y_d}$). i.e.
$\mathcal{L}_{\texttt{reg}} = KL(\mathbf{y_p}||\mathbf{y_d})$

\noindent \textbf{Compatibility Loss ($\mathcal{L}_{\texttt{c}}$).}
ensures that the label distribution ($\mathbf{y_d}$) are not divergent from the original noisy distribution ($\mathbf{\hat{y}}$), i.e. 
$\mathcal{L}_{\texttt{c}} = MSE(\mathbf{y_d},\mathbf{\hat{y}})$

\noindent \textbf{Regularization.}
 $\ell_2$ regularizer is employed to ensure that the network will not stop updating when the predictions become close to the label distribution ($\mathbf{y_d}$). It is defined as ${||\omega||}^{2}_2$, where, $\omega$ is the model parameter.

\noindent \textbf{Overall NLL Loss ($\mathcal{L}_{\texttt{NLL}}$).} The overall loss for NLL module is given by:
$\mathcal{L}_{\texttt{NLL}} = \mathcal{L}_{\texttt{reg}} + \mathcal{L}_\texttt{c} + {||\omega||}^{2}_2$

\subsubsection{Overall MTGLS Loss}
Algorithm~\ref{alg:MTGLS} summarizes the training procedure of the proposed MTGLS. Our overall learning is guided by the following objective functions:

$$\mathcal{L}= \mathcal{L}_{\texttt{pseudo-gaze}} + \mathcal{L}_{\texttt{head-pose}} + \mathcal{L}_{\texttt{{ce}}} + \mathcal{L}_{\texttt{NLL}}$$
Note that, we apply $\mathcal{L}_{\texttt{NLL}}$ on the auxiliary label space of pseudo-gaze and head-pose only as the eye patch is extracted from face using landmark detector is not prone to error (empirically, $\sim$ 100\% accurate). On the other hand, if the pre-trained models~\cite{wang2021facex,albiero2020img2pose} are  more reliable, the dependency on NLL will reduce.

\subsection{Evaluation Protocol of Limited Supervision} \label{sec: ISO}
Our proposed MTGLS learns eye representation from the input eye patches and their pseudo-labels distilled from facial models in terms of pseudo-gaze, head-pose, and eye orientation. Following the standard evaluation protocols, we also adopt Linear Probing (LP) on frozen features~\cite{zhang2016colorful,he2020momentum,caron2021emerging} and Fine-Tuning (FT) the features on downstream tasks~\cite{caron2021emerging} (as illustrated in Fig.~\ref{fig:pipeline} Inference module). For LP, we use data augmentation in terms of random resize crops and horizontal flipping during training phase. Please note that during horizontal flipping the ground truth gaze labels change its sign. For this purpose, weights $\phi$ of the model are frozen and only the parameters of $\mathcal{F}_{\theta}$ are updated. During this gaze adaptation training process, the dataset and task specific samples $\mathbf{x'} \in \mathcal{D}$ are used to adjust the output manifold (i.e. respective label spaces). This adjustment process is enforced by applying the appropriate loss function,  
i.e., gaze direction loss ($\mathcal{L}_{\texttt{gaze}}$) when the output space is a 3-D gaze vector or cross-entropy loss ($\mathcal{L}_{\texttt{gaze-zone}}$) when the output space is a gaze zone, where $\mathcal{L}_{\texttt{gaze}} = \frac{\mathbf{y}}{||\mathbf{y}||_2}. \frac{\mathbf{y'}}{||\mathbf{y'}||_2}$ and $\mathcal{L}_{\texttt{gaze-zone}} = CE(\mathbf{y},\mathbf{y'})$ given $\mathbf{y}$ and $\mathbf{y'}$ are original and predicted labels. 
The overall linear probing process can be expressed as the following optimization problem, where, $\theta_{g} \in \{ \theta_{g_e}, \theta_{g_d}\}$:
$\underset{\mathbf{\phi,\theta_{g}}}{\min} \mathbf{E}_{\mathbf{x',y} \sim \mathcal{D}} [\mathcal{L}_{\texttt{EP}} (\mathcal{F}_\phi \circ  \mathcal{F}_{\theta_{g}}(\mathbf{x'}),\mathbf{y})]$. For FT, the backbone network is initialized with pre-trained weight and it is updated for the downstream task. The downstream task for drivers' gaze zone estimation is a 9-class classification problem. Thus, we additionally evaluate the representative features' quality with a weighted nearest neighbour classifier (k-NN)~\cite{caron2021emerging,wu2018unsupervised}. We freeze the weights of the pre-trained model and extract the penultimate layer's feature for training. The weighted k-NN classifier then performs a similarity matching operation in the feature space to vote for the final label. Our empirical analysis suggests that it worked for 10 NN over several runs. Evaluation paradigm using weighted k-NN is simple and it does not require any complicated setup such as hyperparameter tuning, data augmentation etc. 

\begin{algorithm}[tb]
\caption{Training Procedure for MTGLS}
\label{alg:MTGLS}
\begin{algorithmic}[1]
\Require{$\mathcal{F}_\phi$, $\mathcal{F}_{\theta_g}$, $\mathcal{F}_{\theta_h}$, $\mathcal{F}_{\theta_{lr}}$, and $\mathcal{D}$}
\For{\texttt{$\mathbf{e}$ epochs}} \Comment{MTGLS Training}
\State $\mathbf{g} \gets \texttt{LOS}(\mathbf{x})$ \Comment{Localize Facial Landmarks}
\State $\mathbf{h} \gets \texttt{img2pose}(\mathbf{x})$ \Comment{Estimate head-pose}
\State $\mathbf{lr}\gets \texttt{process}(\mathbf{x})$ \Comment{Get left-right patch}
\State $\mathbf{z}\gets \mathcal{F}_{\phi}(\mathbf{x})$
\State $\mathbf{g'}\gets \mathcal{F}_{\theta_g}(\mathbf{z})$, $h'\gets \mathcal{F}_{\theta_h}(\mathbf{z})$, and $\mathbf{lr'} \gets \mathcal{F}_{\theta_{lr}}(\mathbf{z})$
\State $\mathcal{L}_0 = \mathcal{L}_{\texttt{pseudo-gaze}} (\mathbf{g}, \mathbf{g'}) + \mathcal{L}_{\texttt{head-pose}} (\mathbf{h}, \mathbf{h'}) + \mathcal{L}_{\texttt{ce}} (\mathbf{lr}, \mathbf{lr'}) + \mathcal{L}_{\texttt{NLL}} (\mathbf{g}, \mathbf{g'}) + \mathcal{L}_{\texttt{NLL}} (\mathbf{h}, \mathbf{h'})$
\State $\{\phi, \theta_g, \theta_h, \theta_{lr}\} \gets \triangledown_{\{\phi, \theta_g, \theta_h, \theta_{lr}\}} \mathcal{L}_0$
\EndFor
\For{\texttt{$\mathbf{e}$ epochs}} \Comment{Gaze Estimation}
\State $\mathbf{z} \gets \mathcal{F}_{\phi}(\mathbf{x'})$\Comment{$x' \in \mathcal{D}$}
\State $\mathbf{y'} \gets \mathcal{F}_{\theta_g} (\mathbf{z})$ \Comment{$\theta_g \in \{ \theta_{g_e}, \theta_{g_d}\}$}
\State $\mathcal{L}_1 = \mathcal{L}_{\texttt{LP}} (\mathbf{y}, \mathbf{y'})$  \Comment{$\mathcal{L}_{\texttt{LP}} \in \{\mathcal{L}_{\texttt{gaze}},\mathcal{L}_{\texttt{gaze-zone}}\}$}
\State $\{\theta_{g}\}\gets \triangledown_{\{\theta_{g}\}} \mathcal{L}_1$ \Comment{$\theta_{g} \in \{ \theta_{g_e}, \theta_{g_d}\}$}
\EndFor
\end{algorithmic}
\end{algorithm}

\section{Experiments and Results} \label{sec:exp_res}
We comprehensively compare our method on four publicly available benchmark datasets for different tasks from both quantitative (Sec.~\ref{sec:results}) and qualitative (Sec.~\ref{sec: Qualitative Results}) perspectives. Additionally, we perform extensive ablation studies to investigate the contributions of different individual components of the proposed pipeline (Sec.~\ref{sec: Ablation Studies}).
\subsection{Experimental Protocols}

\noindent \textbf{Datasets.} We evaluate MTGLS on four benchmark datasets: \textbf{CAVE}~\cite{smith2013gaze}, \textbf{MPII}~\cite{zhang2017mpiigaze}, \textbf{Gaze360}~\cite{gaze360_2019}, and \textbf{DGW}~\cite{ghosh2021speak2label}. CAVE~\cite{smith2013gaze} contains 5,880 images of 56 subjects collected in a constrained lab environment with 21 different gaze directions and head-poses for each person. MPII~\cite{zhang2017mpiigaze} gaze dataset has 213,659 images collected from 15 subjects during natural everyday activity in front of a laptop over a three-month duration. Gaze360~\cite{gaze360_2019} is a large-scale dataset collected from 238 subjects in unconstrained indoor and outdoor settings with a wide range of head-poses. DGW~\cite{ghosh2021speak2label} is a large scale driver gaze estimation dataset of 338 subjects collected in an unconstrained `inside a car' scenario during different times of the day. For representation learning, we also used cleaned version of VGGFace2~\cite{cao2018vggface2} containing 3M images of 8,631 identities and cleaned version of MS-Celeb-1M~\cite{guo2016ms} dataset having 1M identities.

\noindent \textbf{Settings.}
To be consistent with the previous methods~\cite{park2018deep,park2019few,yu2019improving,yu2019unsupervised,dubey2019unsupervised}, we follow the same experimental protocol as mentioned in the respective baseline papers for each dataset. The training set for these datasets is used for downstream adaptation and test set for inference. Specifically, we follow `leave-one-person-out' for MPII, cross-validation for CAVE, and Train-Val-Test partitions for the Gaze360 and DGW datasets for our quantitative analysis. For validating few-shot learning, we report the person-specific gaze errors (in \textdegree) for a range of $k$ calibration samples by following the evaluation protocol described in~\cite{park2019few,yu2019improving}.

\noindent \textbf{Implementation Details.}
For images to be in common virtual camera points, we pre-process the input data based on the data-normalization procedure in~\cite{zhang2018revisiting}. For our experiments, we use ResNet-50 as a backbone containing 4 residual blocks with 64, 128, 256 and 512 filters. After training the model $\mathcal{F}$, we freeze the backbone network ($\mathcal{F}_{\phi}$) until the penultimate layer, i.e., the global average pooling layer. As shown in Fig.~\ref{fig:pipeline} (inference module), the feature embedding $\mathbf{z}$ is passed through two FC layers (512 and 256 neurons) for linear probing. The linear probing includes dataset-specific adaptation for gaze estimation and driver gaze zone estimation tasks. The hyperparameters are as follows: we use an SGD optimizer with an initial learning rate of 0.01 having momentum 0.9 and a weight decay of $1 \times e^{-4}$. The batch size is set to 32 with the number of epochs 500. For downstream tasks, we use the SGD optimizer for 20-25 epochs with a learning rate of 0.0001, momentum of 0.9 and weight decay of $ 1 \times e^{-4}$. Further, the value of the hyperparameter K (see Sec.~\ref{sec: NLL}) is tuned on the validation set during MTGLS training and is fixed to 10 as in~\cite{yi2019probabilistic}.

\begin{table}[t]
    \caption{\textbf{Linear Probing (LP) and Fine-Tuning (FT) on CAVE, MPII and Gaze360.} Comparison of different pre-training methods on CAVE, MPII, and Gaze360 datasets. The error is reported in terms of error (in~\textdegree) $\pm$ standard deviation.}
    \label{tab:MTGLS}
    \centering
    \scalebox{0.85}{
    \begin{tabular}{c|c|c|c|c}
    \toprule[0.4mm]
    \rowcolor{mygray} \multicolumn{1}{c|}{\textbf{Method}} & \multicolumn{1}{c|}{\textbf{Pretrain}}  &\multicolumn{1}{c|}{\textbf{Dataset}}    & \multicolumn{2}{c}{\textbf{\begin{tabular}[c]{@{}c@{}}Gaze Error (in \textdegree)\end{tabular}}} \\ \cline{4-5}
    \rowcolor{mygray} \multicolumn{1}{l|}{} &     &    & \multicolumn{1}{c|}{\textbf{LP}}& \multicolumn{1}{c}{\textbf{FT}}\\ \hline \hline
   \multirow{6}{*}{\textbf{MTGLS}}  & VGGFace2 & CAVE & 3.30 $\pm$ 0.97 &  3.16 $\pm$ 0.82\\ 
          & MS-Celeb-1M & CAVE & 3.24 $\pm$ 0.90 & \textbf{3.00 $\pm$ 0.76} \\ 
          & CAVE & CAVE & \textbf{3.20 $\pm$ 1.50} & 3.14 $\pm$ 1.27\\ \cline{2-5}
          & VGGFace2 & MPII & 4.41 $\pm$ 0.98 & 4.23 $\pm$ 0.94\\ 
          & MS-Celeb-1M & MPII & 4.39 $\pm$ 0.89 & \textbf{4.07 $\pm$ 0.95}\\ 
          & MPII & MPII & \textbf{4.21 $\pm$ 1.90} & 4.13 $\pm$ 1.31 \\ \cline{2-5} 
          & VGGFace2 & Gaze360 & 12.90 $\pm$ 3.85 & 12.87 $\pm$ 3.69\\ 
          & MS-Celeb-1M & Gaze360 & 13.00$\pm$ 3.54 & \textbf{12.83 $\pm$ 3.30}\\ 
          & Gaze360 & Gaze360 & \textbf{12.89 $\pm$ 5.72} & 12.88 $\pm$ 5.60 \\ 
    \bottomrule[0.4mm]
    \end{tabular}}
    \vspace{-5mm}
\end{table}
\subsection{Quantitative Results}
\label{sec:results}
\subsubsection{Comparison with Existing Methods}
Following the standard evaluation protocols~\cite{smith2013gaze,zhang2017mpiigaze,gaze360_2019,ghosh2021speak2label}, our empirical results on different datasets are presented in Table~\ref{tab:MTGLS},~\ref{tab:ssl_sota} and~\ref{tab:down_dgw}. In \textbf{Table~\ref{tab:MTGLS}}, we compare the performance of MTGLS pre-trained on different datasets. Among the datasets, CAVE, MPII and Gaze360 have gaze labels; additionally, we explore representation learning from large scale facial datasets i.e. VGGFace2 and MS-Celeb-1M. Please note that the pre-training is performed without using any ground truth gaze labels. We report the results by following traditional evaluation protocol i.e., Linear Probing (LP) and Fine-tuning (FT). The gaze error is presented in error $\pm$ standard deviation format. For LP, pre-training on the same datasets without any label is performing better than the other settings as the downstream distribution is more aligned. However, the representation learned on large scale facial data (i.e. 5M images from MS-Celeb-1M) have less standard deviation in all cases. For FT, the model pre-trained on MS-Celeb-1M is more adaptive as compared to the other settings due to large scale data usage. Similarly in \textbf{Table~\ref{tab:ssl_sota}}, the results suggest that our method consistently achieves competitive performance as compared to the state-of-the-art methods. \textit{Please note that for a fair comparison with the previous works (Table~\ref{tab:ssl_sota}), we only consider the models pre-trained on the same dataset's training partition without their gaze labels.} On CAVE dataset, MTGLS outperforms unsupervised gaze estimation~\cite{yu2019unsupervised} by $\sim$ 6.43\% (3.42\textdegree\ \textrightarrow 3.20\textdegree). Unlike~\cite{yu2019unsupervised}, which adapts data augmentation via gaze redirection learning for unsupervised feature learning, our proposed MTGLS takes advantage of weak-supervision in the form of pseudo-labels obtained from off-the-shelf facial models~\cite{albiero2020img2pose,wang2021facex}. We further note from results in Table~\ref{tab:ssl_sota} that our method even performs better than the few of the supervised state-of-the-art methods on MPII~\cite{park2018deep,wang2019generalizing,cheng2020coarse,FischerECCV2018} when pre-trained on MPII dataset without labels and Gaze360 (13.80\textdegree\ \textrightarrow 12.89\textdegree, $\sim 6.59\%$) datasets. We attribute our gain to the rich knowledge distilled from the models trained on generic face-relevant tasks (i.e. landmark localization~\cite{wang2021facex}, 6-D head-pose estimation~\cite{albiero2020img2pose}). Moreover, MTGLS learns highly generic, and transferable representation as it aligns the auxiliary output manifold with the actual gaze estimation task via multi-task learning.
\begin{table}[t]
\caption{\textbf{Gaze Estimation.} Comparison of different gaze estimation methods on CAVE, MPII, and Gaze360 datasets. \textit{* denotes supervised methods.} MTGLS performs favorably against existing state-of-the-art, and achieves gains upto $\sim 6.43\%$ against unsupervised SOTA on CAVE dataset, and $\sim 6.59\%$ for supervised SOTA on Gaze360 dataset.}
\label{tab:ssl_sota}
\vspace{-2mm}
\begin{center}
\scalebox{0.85}{
\begin{tabular}{c|l|c}
\toprule[0.4mm]
\rowcolor{mygray} \multicolumn{1}{l|}{\textbf{Dataset}} & \begin{tabular}[c]{@{}l@{}}\textbf{Method}\\(Pretrain on same data)  \end{tabular}       & \multicolumn{1}{l}{\textbf{\begin{tabular}[c]{@{}l@{}}Gaze Error (in \textdegree)\end{tabular}}} \\ \hline \hline
\multirow{5}{*}{\textbf{CAVE}}         & Jyoti et al.~\cite{jyoti2018automatic}*         & \textbf{2.22}                                                                                              \\  
                                       & Park et al.~\cite{park2018deep}*             & 3.59                                                                                              \\
                                       & Yu et al.~\cite{yu2019unsupervised}         & 3.42                                                                                              \\  
                                       & MTGLS                     & 3.20 \\
                                       & MTGLS (MS-Celeb-1M)                     & 3.00 \\ \midrule
\multirow{6}{*}{\textbf{MPII}}         & Park et al.~\cite{park2018deep}*             & 4.50                                                                                               \\
 & Wang et al.~\cite{wang2019generalizing}* & 4.30 \\
 & Cheng et al.~\cite{cheng2020coarse}* & 4.10 \\
 & Fischer et al.~\cite{FischerECCV2018}* & 4.30 \\
 & MTGLS                    & 4.21              \\ 
  & MTGLS (MS-Celeb-1M)                    & \textbf{4.07}                                                                                               \\ \midrule
\multirow{3}{*}{\textbf{Gaze360}}      & Kellnhofer et al.~\cite{gaze360_2019}*     & 13.80                                                                                              \\ 
                                       & MTGLS                    & 12.89                                                                                             \\ 
                                       & MTGLS  (MS-Celeb-1M)                  & \textbf{12.83}                                                                                             \\ \bottomrule[0.4mm]
\end{tabular}}
\end{center}
\vspace{-8mm}
\end{table}
\begin{table}[b]
\caption{\textbf{k-NN, Linear Probing (LP) and Fine-Tuning (FT) results on DGW dataset.} Performance comparison with state-of-the-art (SOTA) methods on driver gaze zone estimation on DGW dataset. \textit{* denotes supervised methods.} Note that even after learning with limited supervision, MTGLS performs favorably against most of the supervised methods on DGW dataset.}
\label{tab:down_dgw}
\vspace{-5mm}
\begin{center}
\scalebox{0.85}{
\begin{tabular}{l|c|c}
\toprule[0.4mm]
\rowcolor{mygray}  \textbf{{Method}}& \textbf{\begin{tabular}[c]{@{}c@{}}Validation\\ Accuracy (\%)\end{tabular}} & \textbf{\begin{tabular}[c]{@{}c@{}}Test\\ Accuracy (\%)\end{tabular}} \\ \hline \hline
Vasli et al.~\cite{vasli2016driver}*     &   {52.60}                                                                        &   {50.41}                                                                   \\ \hline
Tawari et al.~\cite{tawari2014driver}*    &   {51.30}                                                                        &   {50.90}                                                                   \\ \hline
Fridman et al.~\cite{fridman2015driver}*     &   {53.10}                                                                        &   {52.87}                                                                   \\ \hline
\begin{tabular}[c]{@{}c@{}}Vora et al.~\cite{vora2017generalizing} (Alexnet face)* \end{tabular} &  {56.25  }                                                                         &  {57.98 }                                                                   \\ \hline
\begin{tabular}[c]{@{}c@{}}Vora et al.~\cite{vora2017generalizing} (VGG face)*\end{tabular}     &  {58.67}                                                                        &  {58.90 }                                                                   \\ \hline
SqueezeNet~\cite{iandola2016squeezenet}*                                                                             &  {59.53}                                                                           &  {59.18 }                                                                    \\ \hline
Ghosh et al.~\cite{ghosh2021speak2label}*                                                                              &  {60.10}                                                                           &  {60.98}                                                                    \\ \hline
Inception V3~\cite{DBLP:journals/corr/SzegedyVISW15}*                                                                           &  {67.93}                                                                           &  {68.04}                                                                    \\ \hline
Vora et al.~\cite{vora2018driver}*                                                         &  { 67.31}                                                       &  {68.12}                                                \\ \hline
ResNet-152~\cite{DBLP:journals/corr/HeZRS15}*                                                                              &  {68.94}                                                                           &  {69.01 }                                                                   \\ \hline
\begin{tabular}[c]{@{}c@{}}Yoon et al.~\cite{yoon2019driver} (Face + Eyes)* \end{tabular}     &   {70.94}                                                                        &   {71.20}                                                                   \\ \hline

\begin{tabular}[c]{@{}c@{}}Stappen et al.~\cite{stappen2020x}*\\\end{tabular}     &   {71.03}                                                                        &   {71.28}                                                                   \\\hline
\begin{tabular}[c]{@{}c@{}}Lyu et al.~\cite{lyu2020extract}*\\\end{tabular}     &   {85.40}                                                                        &   {81.51}                                                                   \\ \hline
\begin{tabular}[c]{@{}c@{}}Yu et al.~\cite{yu2020multi}* \\\end{tabular}     &   {80.29}                                                                        &   {82.52}                                                                   \\\hline
\begin{tabular}[c]{@{}c@{}}MTGLS (k-NN)\end{tabular}                                                                   & 63.90                    & 65.35             \\\hline
\begin{tabular}[c]{@{}c@{}}MTGLS (LP)\end{tabular}                                                                   & 73.83                    & 75.10              \\\hline
MTGLS (FT)                                                                  & 79.63                   & 80.05           \\ \bottomrule[0.4mm]
\end{tabular}}
\end{center}
\vspace{-8mm}
\end{table}
\begin{figure*}[t]
    \centering
    \includegraphics[width=0.32\linewidth]{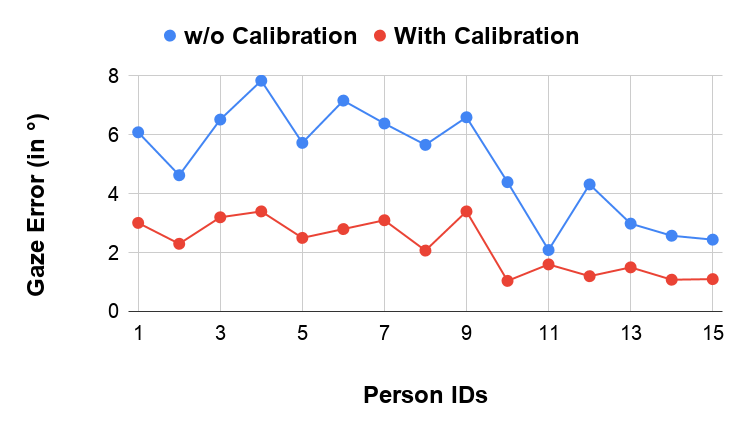}
    \includegraphics[width=0.32\linewidth]{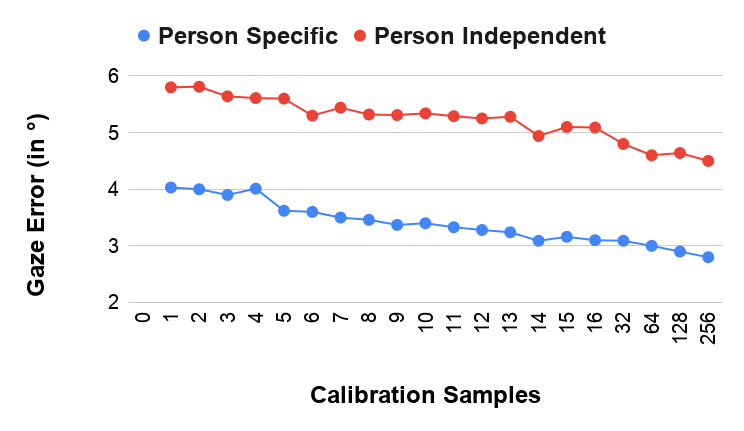}
    \includegraphics[width=0.32\linewidth]{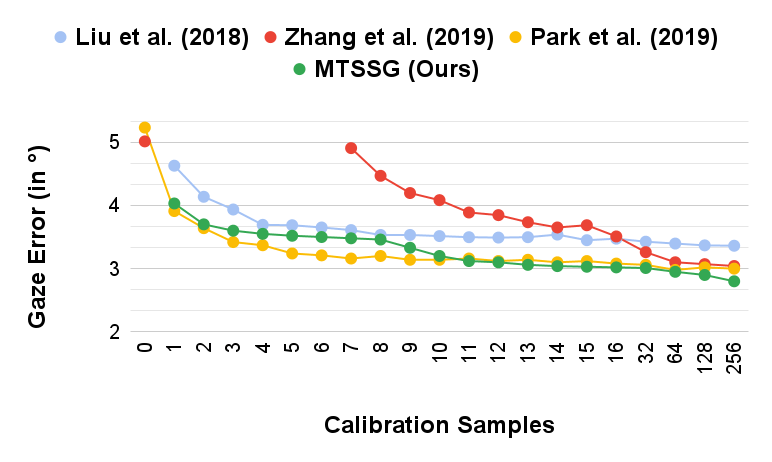}
    \caption{\textbf{Left}: Comparison of with and without person-specific calibration on MPII dataset. \textbf{Middle}: Comparison with person-specific and person-independent setting. \textbf{Right}: Comparison with state-of-the-art few-shot learning approaches. (Refer Sec.~\ref{sec:person})}
    \label{fig:person}
    \vspace{-5mm}
\end{figure*}
Further, we evaluate MTGLS on a different yet relevant task of driver gaze zone estimation on DGW dataset~\cite{ghosh2021speak2label} to show the adaptability of MTGLS across tasks. The results are shown in Table~\ref{tab:down_dgw}. Please note that the compared methods in Table~\ref{tab:down_dgw} are supervised. MTGLS performs favorably against~\cite{vasli2016driver, tawari2014driver,fridman2015driver,vora2018driver,vora2017generalizing,yu2020multi,lyu2020extract}. Some of the compared methods~\cite{yu2020multi,lyu2020extract} use additional information in the form of body pose and other cues for driver gaze estimation which are more aligned towards the driver monitoring task. Our simple eye patch-based representation learning (i.e. MTGLS (FT)) outperforms the baseline~\cite{ghosh2021speak2label} by $19.53\%$ and $19.08\%$ margin on validation and test set, respectively. Even with weighted k-NN on the features of the penultimate layer, our method performs better than the supervised baseline~\cite{ghosh2021speak2label} ($3.80\%$ on the validation set and $4.37\%$ on the test set). These results indicate that MTGLS learns a highly generic and transferable eye representation using complementary auxiliary tasks.

\subsubsection{Person Specific and Few-Shot Gaze Estimation} \label{sec:person}
Person-specific adaptation can further improve gaze estimation performance and has therefore gained a lot of research attention~\cite{yu2019improving, park2019few,swaminathan2018enabling,blattgerste2018advantages} due to its potential applications in AR/VR industry. 

\noindent \textbf{With and w/o Person Specific Calibration [Fig.~\ref{fig:person} (Left)].}
Similar to~\cite{park2019few,yu2019improving}, we conduct person-specific calibration on the MPII dataset having 15 identities. We randomly choose 200 calibration samples for each identity for adaptation and evaluate the performance on the remaining samples. The results in terms of gaze error (in~\textdegree) corresponding to different person IDs are shown in Fig.~\ref{fig:person} (Left). The results suggest that the angular error reduces significantly with the calibration. The average angular error reduces to 2.22\textdegree\ (with calibration) from 5.02\textdegree\ (without calibration). Please note that this average is not plotted in Fig.~\ref{fig:person} (Left).

\noindent \textbf{Person specific vs. Person Independent Calibration [Fig.~\ref{fig:person} (Middle)].}
Apart from person-specific adaptation, we perform person-independent adaptation to show the generalization capabilities of the MTGLS inspired by~\cite{park2019few}. The underlying hypothesis behind this is that the latent representation is different across subjects due to the individuality of the eyes. For the person independent setting, we randomly choose 1-256 calibration samples for adaptation and evaluate the performance on the disjoint test set. For person dependent setting, we randomly choose 1-256 calibration samples for adaptation and evaluate the performance on the disjoint remaining person-specific test set. We report the average over the identities in the figure. The results are depicted in Fig.~\ref{fig:person} (Middle), representing the number of calibration samples (x-axis) and corresponding gaze error \textdegree\ (y-axis). From these results, we observe that for all of the calibration points varying from 1-256, the person-specific adaptation performs better than the person-independent setting. Another intuitive trade-off observed from the graph is that the difference between person-independent and person-specific adaptation decreases, as the number of samples increases. For example, between 1 and 256 samples the difference between two settings reduces by $\sim 7\%$ (1 sample: 37.77\% [5.80\textdegree\ \textrightarrow 4.03\textdegree], 256 samples: 30.51\% [4.50\textdegree\ \textrightarrow 2.8\textdegree]). This experiment further demonstrates the capabilities of MTGLS to learn generic representation. 

\noindent \textbf{Few-Shot Gaze Estimation [Fig.~\ref{fig:person} (Right)].}
Few-shot person-specific gaze estimation has recently gained popularity~\cite{park2019few,yu2019improving}, as it enables person-specific gaze calibration using only a few samples. Here, we investigate the transferable characteristics of our features for few-shot gaze estimation. We follow the same evaluation protocol as in~\cite{park2019few,yu2019improving} for a fair and direct comparison with existing methods. Among 15 subjects of the MPII dataset, the last 500 images are reserved for testing, while $k$ calibration samples are randomly selected for adaptation purposes. The results are shown in Fig.~\ref{fig:person} (Right) indicate that our method performs favorably against the existing works (for $\geq$ 12), thus demonstrating the strength of our learned representations for low-annotation data regimes. The results for compared methods are adapted from~\cite{park2019few}.


\begin{table}[b]
\caption{Contributions of different auxiliary tasks, their combinations, and noise correction towards overall MTGLS performance. Results are presented in terms of gaze error (in \textdegree) on CAVE dataset.}
\label{tab:ablation_auxiliary}
\begin{center}
\scalebox{0.85}{
\begin{tabular}{c|c|c|c|c}
\toprule[0.4mm]
\rowcolor{mygray} \textbf{Pseudo-Gaze} & \textbf{Head-Pose} & \textbf{L/R Eye} & \textbf{NLL} & \textbf{\begin{tabular}[c]{@{}l@{}} Error (in \textdegree)\end{tabular}} \\ \hline  \hline
$\checkmark$                 &                   &  &                         & 3.40                  \\ \hline
                     & $\checkmark$              &       &                   & 3.52                 \\ \hline
                     &                   & $\checkmark$          &           & 3.67                 \\ \hline
$\checkmark$                 & $\checkmark$              &                   &       &          3.30            \\ \hline
                     & $\checkmark$              & $\checkmark$                    &            &    3.31       \\ \hline
$\checkmark$                 &                   & $\checkmark$                    &                &   3.33    \\ \hline
$\checkmark$                 & $\checkmark$              & $\checkmark$               &      & 3.23                 \\ \hline
$\checkmark$                 & $\checkmark$              & $\checkmark$                    & $\checkmark$   &  3.00                 \\ \bottomrule[0.4mm]
\end{tabular}}
\end{center}
\vspace{-5mm}
\end{table}


\subsubsection{Ablation Studies} \label{sec: Ablation Studies}

\noindent \textbf{Contribution of the Auxiliary Tasks.}
To study the impact of different auxiliary tasks on the overall performance of MTGLS, we progressively integrate them into our framework by keeping other parts fixed. We assess the impact of each auxiliary task and their different combinations during the representation learning on the CAVE dataset~\cite{smith2013gaze}. From the results in Table~\ref{tab:ablation_auxiliary}, it can be observed that the contribution of head-pose (3.52\textdegree) and pseudo eye gaze (3.40\textdegree) is greater as compared to the eye orientation (3.67\textdegree), indicating that head-pose and pseudo-gaze provide more relevant information for gaze estimation as compared to eye orientation. Similarly, when the model is trained with both head-pose and pseudo-gaze, the effect is less as compared with the other combinations. Remarkably, the best performance is achieved when all of the three auxiliary tasks are combined i.e., by 2.12\% (3.30\textdegree\ \textrightarrow 3.23\textdegree). These empirical evaluations establish the complimentary strength and individual importance of each of the proposed auxiliary tasks in our framework.

\noindent \textbf{Contribution of the Noise Removal in Label Space.}
Since two of our supervisory signals (pseudo-gaze and 6-D head-pose labels) contain inherent noise, MTGLS incorporates a noise correction strategy (see Sec.~\ref{sec: NLL}) based upon label distribution modelling. Our experiments reveal that when we integrate the NLL framework into MTGLS, the performance is significantly enhanced $ \sim 7.12\%$ (3.23\textdegree\ \textrightarrow 3.00\textdegree) thus suggesting the importance of this module. 

\noindent \textbf{Choice of Landmark Localization Model.} Two of our auxiliary learning signals come from off-the-shelf face analysis models developed for landmark localization~\cite{wang2021facex} and head-pose estimation~\cite{albiero2020img2pose}. Here, we evaluate the impact on the performance of MTGLS by considering other teacher models for localizing facial landmarks for gaze estimation. Our experiments on MPII dataset reveal that our landmark guided pseudo gaze method is more accurate than OpenFace 2.0 tracker~\cite{baltrusaitis2018openface} ($\sim$ 4.39\% i.e. 9.10\textdegree\ \textrightarrow 8.7\textdegree) and  OpenFace tracker~\cite{baltruvsaitis2016openface} ($\sim$ 12.65\% i.e. 9.96\textdegree\ \textrightarrow 8.7\textdegree). These empirical evaluations justify our choice of~\cite{wang2021facex} for landmark localization.

\begin{figure} [t]
    \centering
    \includegraphics[width=\linewidth, height =5cm]{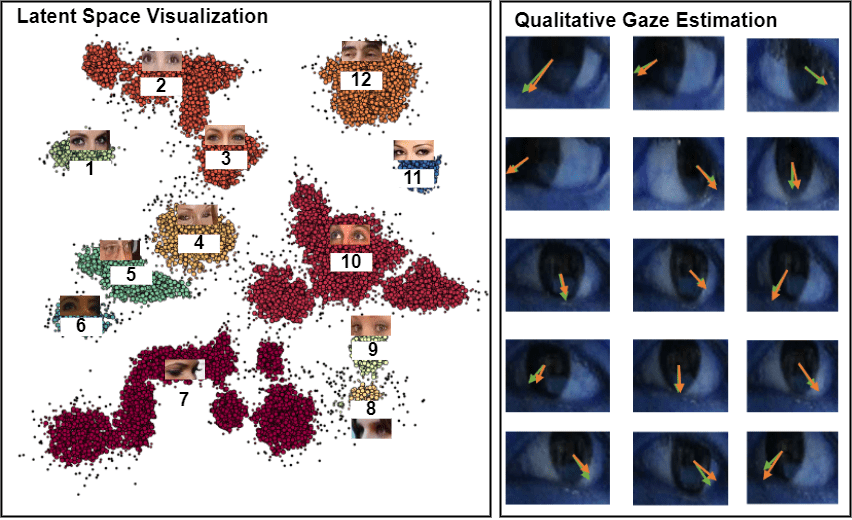}
    \caption{\textbf{Left:} t-SNE visualization of the learned feature space by MTGLS. We observe well-formed clusters corresponding to different gaze directions and head-poses \textbf{Right:} Qualitative prediction result on CAVE dataset. Here, orange and green arrows represent predicted and ground-truth gaze direction, respectively.}
    \label{fig:tsne}
    \vspace{-5mm}
\end{figure}

\subsection{Qualitative Results}\label{sec: Qualitative Results}
\noindent \textbf{Visualization of Gaze Representation.}
To further understand the effectiveness of learned latent representation qualitatively, we compute t-SNE~\cite{van2008visualizing} on the learned features space. After mapping the high dimensional feature to 2-D, DBSCAN clustering is performed (with the maximum distance = 1 and neighbourhood minimum sample = 12). The output is shown in Figure~\ref{fig:tsne}. We observe that the 2-D mapping of latent distributions is mapped to overlapping clusters. We further overlay a representative eye patch corresponding to each of the formed clusters and observe that each cluster represents different head-pose, gaze, and their combination. We can notice clear patterns in learned space, i.e., the clusters within close proximity of each other have visual similarities in terms of eye gaze and head-pose (e.g., cluster pairs 2-3, 8-9, and 11-12). 

\noindent \textbf{Gaze Estimation.} Fig.~\ref{fig:tsne} (Right) illustrates the quality of our gaze estimation results. Here, the orange and green arrows indicate predicted and ground-truth gaze direction, respectively. It reflects the changes in the gaze estimation output manifold after representation learning via the proposed MTGLS. From the image, it is observed that our framework learns robust gaze estimation manifold. 

\section{Conclusion} \label{sec:conclusion}

We have proposed a multi-task gaze representation learning framework with limited supervision that exploits different auxiliary signals to minimize the misalignment between the output manifold of the actual and auxiliary tasks. Our method learns a robust feature embedding from abundantly available non-annotated facial images by distilling knowledge from deep CNN models developed for other facial analysis tasks. To counter the unavoidable noise in the auxiliary label space, we incorporate a probabilistic label distribution matching approach. Through a set of extensive empirical evaluations and ablations, we demonstrate that the gaze representations learned by our method are robust to several factors including appearance, head-poses, and illumination, and generalize well across different tasks and datasets. We further show that our generic representations transfer well for person-specific and few-shot calibrations. In future, it would be interesting to explore design choices across multi-task learning approaches, optimal pre-training strategies and combining the multi-task losses.

{\small
\bibliographystyle{ieee_fullname}
\bibliography{MTGLS}
}
\end{document}